\documentclass[10pt,twocolumn,letterpaper]{article}

\usepackage[pagenumbers]{cvpr} 

\usepackage{makecell}
\usepackage{float}

\usepackage[dvipsnames]{xcolor}

\definecolor{cvprblue}{rgb}{0.21,0.49,0.74}
\usepackage[pagebackref,breaklinks,colorlinks,citecolor=cvprblue]{hyperref}

\def\paperID{15006} 
\def\confName{CVPR}
\def\confYear{2024}

\title{Mitigating Exposure Bias in Discriminator Guided Diffusion Models}

\author{Eleftherios Tsonis, Paraskevi Tzouveli, Athanasios Voulodimos\\ \\
Artificial Intelligence and Learning Systems Laboratory\\
School of Electrical and Computer Engineering\\
National Technical University of Athens
}

\begin{document}
\maketitle
\begin{abstract}
Diffusion Models have demonstrated remarkable performance in image generation. However, their demanding computational requirements for training have prompted ongoing efforts to enhance the quality of generated images through modifications in the sampling process. A recent approach, known as Discriminator Guidance, seeks to bridge the gap between the model score and the data score by incorporating an auxiliary term, derived from a discriminator network. We show that despite significantly improving sample quality, this technique has not resolved the persistent issue of Exposure Bias and we propose SEDM-G++, which incorporates a modified sampling approach, combining Discriminator Guidance and Epsilon Scaling. Our proposed approach outperforms the current state-of-the-art, by achieving an FID score of 1.73 on the unconditional CIFAR-10 dataset. 
\end{abstract}    
\section{Introduction}
\label{sec:intro}
Diffusion models, initially proposed by Sohl-Dickstein \etal in 2015 \cite{sohldickstein2015deep}, have excelled in various domains, including image generation, audio generation \cite{kong2020diffwave, chen2020wavegrad} and video generation \cite{ho2022video, voleti2022video, blattmann2023align}. In the domain of image synthesis, diffusion models have seen significant advancements in subsequent years through works such as those of Song and Ermon (2019) \cite{song2019generative}, Ho \etal (2020) \cite{ho2020ddpm}, and Nichol and Dhariwal (2021) \cite{nichol2020improvedddpm}. In 2021, Song \etal presented a novel approach that unifies score-based models and Denoising Diffusion Probabilistic Models (DDPMs) by employing stochastic differential equations (SDEs) \cite{song2021sde}. Moreover, Karras \etal (2022) presented a comprehensive exploration of the diffusion model design space and introduced the EDM model \cite{karras2022edm}, which implemented a range of optimizations to both the sampling and training processes, leading to a substantial enhancement in performance and sample quality.

\begin{figure}[ht]
    \centering
    \includegraphics[width=\linewidth]{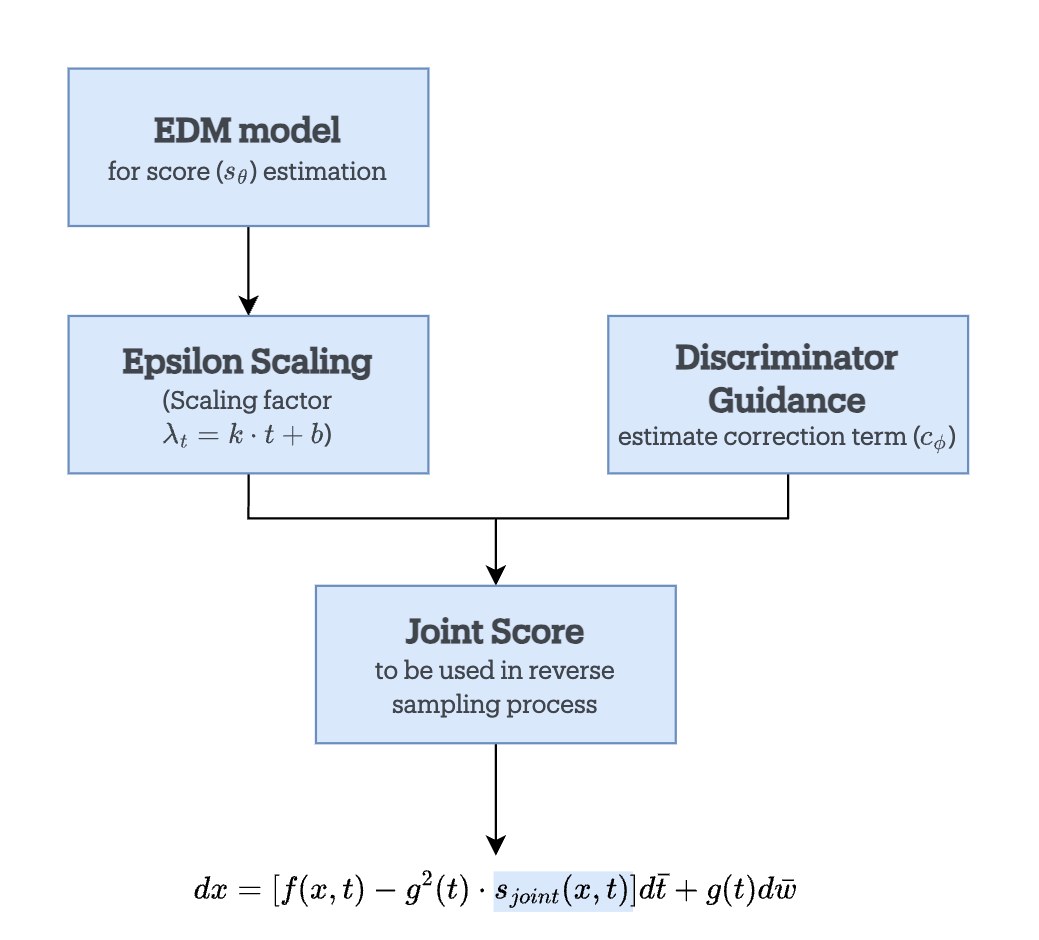}
    \caption{Overview of our proposed SEDM-G++.}
    \label{fig:short}
\end{figure}

The computational cost associated with training new score models from the ground up has initiated research endeavors which employ pre-existing score models and enhance the quality of generated samples via refinements in the sampling procedure.

In an effort loosely inspired by Generative Adversarial Networks (GANs), Kim \etal \cite{kim2023refining} modify the diffusion sampling process by utilising a discriminator network to bridge the gap between the model score and the true data score. In their method, \textit{Discriminator Guidance} (DG), they maintain the pre-trained score model as a fixed component and introduce a discriminator network to classify real and generated data across different noise scales. The method incorporates a correction term into the model score, guided by the discriminator's feedback, which helps steer the sample generation towards more realistic paths.

The iterative sampling chain in diffusion models is long, usually requiring thousands of steps due to the Gaussian assumption of reverse diffusion, which only holds for small step sizes \cite{sohldickstein2015deep}. This leads to the exposure bias problem, illustrated by Ning \etal \cite{ning2023a}. Exposure bias refers to the discrepancy between the input data during training and inference phases. During training, the model is consistently exposed to the ground truth training sample $\pmb{x}_t$. However, during inference, the model relies on the previously generated sample, $\hat{\pmb{x}}_t$. This distinction between $\pmb{x}_t$ and $\hat{\pmb{x}}_t$ results in a difference between $\pmb{\epsilon}_{\pmb{\theta}}(\pmb{x}_t)$ and $\pmb{\epsilon}_{\pmb{\theta}}(\hat{\pmb{x}}_t)$, where $\pmb{\epsilon}_{\pmb{\theta}}$ is the model's noise prediction. This disparity between the two predictions results in error accumulation and deviations in the sampling process, known as "sampling drift" \cite{li2023alleviating}. Ning \etal \cite{ning2023b} propose an effective method, \textit{Epsilon Scaling} for alleviating exposure bias, which is incorporated directly into the sampling process and requires no training or fine-tuning of the model.

The preceding research prompts us to inquire whether Discriminator Guidance is effective in mitigating the accumulation of exposure bias in the sampling process. Our findings indicate that, despite notable enhancements in sample quality, Discriminator Guidance is ineffective in alleviating exposure bias in Diffusion Models. We propose SEDM-G++, which incorporates a modified sampling approach, combining Discriminator Guidance and Epsilon Scaling. Fig. \ref{fig:short} shows an overview of SEDM-G++. We test our method on top of the pre-trained EDM model \cite{karras2022edm} and show that the proposed sampling process achieves improved sample quality, while reducing exposure bias.

Our contributions can be summarized as follows.
\begin{itemize}
    \item We investigate exposure bias in discriminator guided diffusion models.
    \item We propose SEDM-G++, which incorporates a sampling approach combining Discriminator Guidance and Epsilon Scaling.
    \item Our proposed method
    improves sample quality across the board and outperforms the current state-of-the-art, by achieving an FID score of 1.73 on the unconditional CIFAR-10 dataset.
\end{itemize}
We intend to make the code publicly available, upon publication.
\section{Related Work}
\label{sec:related_work}

Diffusion models \cite{sohldickstein2015deep} have demonstrated exceptional performance across diverse domains such as image, audio, and video generation \cite{kong2020diffwave, chen2020wavegrad, ho2022video, voleti2022video, blattmann2023align}. In the realm of image synthesis, substantial progress has been made in recent years through various contributions \cite{song2019generative, song2020denoising, ho2020ddpm, nichol2020improvedddpm, song2021sde, karras2022edm, rombach2022high, meng2023distillation, jing2022subspace, bond2022unleashing}. Diffusion models find a wide range of applications, including text to image generation \cite{ruiz2023dreambooth, gu2022vector, liu2022compositional}, image inpainting \cite{rombach2022high, lugmayr2022repaint, chung2022come, wang2023imagen}, image editing \cite{kawar2023imagic, avrahami2022blended, kim2022diffusionclip, mokady2023null, preechakul2022diffusion, chung2022come, choi2021ilvr}, super-resolution \cite{chung2022come, gao2023implicit}, point cloud generation \cite{luo2021diffusion}, 3D shape generation \cite{zhou20213d} and vision decoding \cite{chen2023seeing}.

Ning \etal \cite{ning2023a} identify a phenomenon associated with the sampling chain, which involves the accumulation of errors across $T$ inference sampling steps. This accumulation is primarily attributed to the discrepancy between the training and inference stages. During training, the diffusion model is trained with a ground truth pair $(\pmb{x}_t, \pmb{x}_{t-1})$, learning to reconstruct $\pmb{x}_{t-1}$ given $\pmb{x}_t$. However, during inference, the model lacks access to the ground truth $\pmb{x}_t$ and relies on the previously generated $\hat{\pmb{x}}_t$, leading to a potential accumulation of errors. This mismatch between the input used in training and the input used in sampling resembles the exposure bias problem, originally observed in other generative models \cite{schmidt2019generalization, ranzato2016sequence}.

To address the exposure bias issue, Ning \etal \cite{ning2023a} suggest explicitly modeling the prediction error during training. During the training phase, they perturb $\pmb{x}_t$ as normal and provide the network with a new, noisier version of $\pmb{x}_t$. This simulates the training-sampling discrepancy, fooling the learned network into considering potential prediction errors during inference. Even though their method proves effective in reducing the exposure bias phenomenon, it is cumbersome as it necessitates retraining the score network entirely, a computationally expensive endeavor.

Li \etal \cite{li2023alleviating} propose a different approach, which involves shifting the timestep $t$ during sampling. They observe that the time step $t$ is directly linked to the corruption level of data samples, and demonstrate that adjusting the subsequent time step $t-1$ during sampling, based on the variance of the currently generated samples, can effectively mitigate exposure bias. Despite the fact that their method circumvents the need for model retraining, tuning the timestep shift is difficult to optimize.
\section{Background}
\subsection{Diffusion Models Framework}

In DDPMs, Ho \etal \cite{ho2020ddpm} define the forward diffusion process as the gradual addition of noise to a clean datum, until structure is destroyed entirely and the datum is transformed into pure noise. Consider the data distribution $\pmb{x}_0\sim q(\pmb{x}_0)$. The forward process is essentially a Markov chain, which generates a sequence of random variables $\pmb{x}_1,\pmb{x}_2,\dots \pmb{x}_T$ with transition kernel $q(\pmb{x}_t|\pmb{x}_{t-1})$. A variance schedule $\{\beta_t\in (0,1)\}_{t=0}^T$, such that the noise perturbing the data at each discrete timestep $t$ is an isotropic Gaussian with variance $\beta_t$, is inherent to the DDPM framework. The Markov chain is defined by the transition kernel:
\begin{equation}
    q(\pmb{x}_t | \pmb{x}_{t-1}) = {\cal N}(\pmb{x}_t; \sqrt{1- \beta _t} \pmb{x}_{t-1}, \beta _t\pmb{I})
    \label{eq:dif1}
\end{equation}
The joint distribution of $\pmb{x}_1,\pmb{x}_2,\dots \pmb{x}_T$, conditioned on $\pmb{x}_0$ is:
\begin{equation}
    q(\pmb{x}_{1:T}|\pmb{x}_0)=\prod_{t=1}^T q(\pmb{x}_t|\pmb{x}_{t-1})
    \label{frwd_joint}
\end{equation}

A significant quality of the forward diffusion process is that $\pmb{x}_t$, drawn from $q(\pmb{x}_t|\pmb{x}_{t-1})$ at timestep $t$, can be directly expressed as a linear combination of $\pmb{x}_0$ and a noise variable $\pmb{\epsilon}\sim {\cal N} (\pmb{0}, \pmb{I})$:
\begin{equation}
    \pmb{x}_t=\sqrt{\bar{a}}_t\pmb{x}_0+\sqrt{1-\bar{a}_t}\pmb{\epsilon}
    \label{xt_x0_cond}
\end{equation}
where $\bar{a}_t=\prod_{i=1}^t (1-\beta_i)$ and $\alpha_t=1-\beta_t$.

For the reverse process, seeing as estimating $q(\pmb{x}_{t-1}|\pmb{x}_t)$ is unfeasible, a learnable transition kernel $p_{\pmb{\theta}}$ is utilised instead, to approximate the conditional probabilities. The kernel takes the following form:
\begin{equation}
    p_{\pmb{\theta}}(\pmb{x}_{t-1}|\pmb{x}_t)=\mathcal{N}(\pmb{x}_{t-1};\mu_{\pmb{\theta}}(\pmb{x}_t,t), \Sigma_{\pmb{\theta}}(\pmb{x}_t,t))
    \label{rev_kernel}
\end{equation}
where $\pmb{\theta}$ denotes model parameters. By sampling ${\pmb{x}_T \sim {\cal N} (\pmb{0}, \pmb{I})}$ and iteratively using Eq. \ref{rev_kernel} to run the reverse diffusion process, we obtain a sample from $q(\pmb{x}_0)$.

The success of the sampling process depends on training the reverse Markov chain to match the time reversal of the forward Markov chain. This involves adjusting parameter $\pmb{\theta}$ to make the joint distribution of the reverse Markov chain $p_{\pmb{\theta}}(\pmb{x}_{0:T})=p(\pmb{x}_T) \prod_{t=1}^T p_{\pmb{\theta}}(\pmb{x}_{t-1}|\pmb{x}_t)$ closely approximate the joint distribution of the forward process $q(\pmb{x}_{0:T})=q(\pmb{x}_0) \prod_{t=1}^T q(\pmb{x}_t|\pmb{x}_{t-1})$ (Eq. \ref{frwd_joint}). This is accomplished by minimizing the Kullback-Leibler (KL) divergence between $p_{\pmb{\theta}}(\pmb{x}_{t-1}|\pmb{x}_t)$ and $q(\pmb{x}_{t-1}|\pmb{x}_t, \pmb{x}_0)$:
\begin{equation}
    D_{KL}(q(\pmb{x}_{t-1}|\pmb{x}_t, \pmb{x}_0 \,||\, p_{\pmb{\theta}}(\pmb{x}_{t-1}|\pmb{x}_t))
\end{equation}
Ho \etal show that it is possible to directly compare $p_{\pmb{\theta}}(\pmb{x}_{t-1}|\pmb{x}_t)$ against forward process posteriors, which are tractable when conditioned on $\pmb{x}_0$ \cite{ho2020ddpm}:
\begin{equation}
    q(\pmb{x}_{t-1}|\pmb{x}_t,\pmb{x}_0) =  {\cal N}(\pmb{x}_{t-1}; \tilde{\pmb{\mu}_t}(\pmb{x}_t, \pmb{x}_0), \tilde\beta_t \pmb{I}),
\end{equation}
where:
\begin{equation}
    \tilde{\pmb{\mu}_t}(\pmb{x}_t, \pmb{x}_0) = \frac{\sqrt{\bar\alpha_{t-1}}\beta_t }{1-\bar\alpha_t}\pmb{x}_0 + \frac{\sqrt{\alpha_t}(1- \bar\alpha_{t-1})}{1-\bar\alpha_t} \pmb{x}_t
    \label{mu_t}
\end{equation}
\begin{equation}
    \tilde\beta_t = \frac{1-\bar\alpha_{t-1}}{1-\bar\alpha_t}\beta_t  
\end{equation}

By plugging Eq. \ref{xt_x0_cond} into Eq. \ref{mu_t} we derive that given $\pmb{x}_t$, $\mu_{\pmb{\theta}}$ must predict the following quantity:
\begin{equation}
    \mu_{\pmb{\theta}}(\pmb{x}_t,t) = \frac{1}{\sqrt{a_t}}\left(\pmb{x}_t-\frac{\beta_t}{\sqrt{1-\bar a_t}} \pmb{\epsilon_\theta} (\pmb{x}_t,t)\right)
    \label{eq:mu_sample}
\end{equation}
where $\pmb{\epsilon_\theta}$ is the approximate prediction for noise factor $\pmb{\epsilon}$ based on $\pmb{x}_t$.

\subsection{Stochastic Differential Equations}
Song \etal \cite{song2021sde} demonstrated that DDPMs can be regarded as discretizations of stochastic differential equations (SDEs), thus providing a unifying perspective on previous approaches. The forward diffusion process is described by the forward SDE:
\begin{equation}
\label{eq:forward_sde}
d\pmb{x}=\pmb{f}(\pmb{x},t)dt+g(t)d \pmb{w},
\end{equation}
where $\pmb{f}(\pmb{x},t)$ is the drift coefficient, $g(t)$ is the diffusion coefficient and $t$ is a continuous time variable in $[0,T]$.

Anderson \cite{anderson1982reverse} demonstrated that reversing a diffusion process yields another diffusion process, which operates backward in time and is described by the reverse-time stochastic differential equation:
\begin{equation}
\label{eq:reverse_sde}
    d\pmb{x}=\big[\pmb{f}(\pmb{x},t)-g^{2}(t)\nabla\log{q(\pmb{x})}\big]d\bar{t}+g(t)d\bar{\pmb{w}},
\end{equation}
where $d\bar{t}$ and $\bar{\pmb{w}}$ denote the infinitesimal reverse-time and reverse-time Brownian motion, respectively. Consequently, the continuous-time generative process is formulated as:
\begin{equation}
    d\pmb{x}=\big[\pmb{f}(\pmb{x},t)-g^{2}(t)\pmb{s}_{\pmb{\theta}}(\pmb{x},t)\big]d\bar{t}+g(t)d\bar{\pmb{w}},
\end{equation}
Here, the score network's estimation target $\pmb{s}_{\pmb{\theta}}(\pmb{x},t)$ corresponds to the actual data score $\nabla\log{q(\pmb{x})}$.

\subsection{Discriminator Guidance}
We investigate Exposure Bias in the discriminator guided \textit{EDM-G++} (Kim \etal \cite{kim2023refining}). Their approach keeps the score function frozen and trains a discriminator network in isolation. The absence of adversarial training makes convergence simpler and faster to achieve, compared to GANs.
When producing samples using the reverse-time SDE:
\begin{equation}
    d\pmb{x} = [\pmb{f}(\pmb{x}, t) - g(t)^2  \pmb{s}_{\pmb{\theta}_\infty}(\pmb{x},t)] dt + g(t)d \bar{\pmb{w}}
\end{equation}
Kim \etal \cite{kim2023refining} show that the generative process might diverge from the reverse-time data process if the local optimum $\pmb{\theta}_\infty$ of the score network $\pmb{s}_{\pmb{\theta}_\infty}$ is different to the global optimum $\pmb{\theta}_*$. However, augmenting the score function by a \textit{correction term} can bridge the gap between the two processes.

The reverse-time SDE with the adjusted score becomes:
\begin{equation}
    d\pmb{x} = [\pmb{f}(\pmb{x}, t) - g(t)^2  (\pmb{s}_{\pmb{\theta}_\infty} + \pmb{c}_{\pmb{\theta}_\infty})(\pmb{x},t)] dt + g(t)d \bar{\pmb{w}}
\end{equation}
for $\pmb{c}_{\pmb{\theta}_\infty}(\pmb{x},t) = \nabla \log \frac{p_r^t(\pmb{x})}{p^t_{\pmb{\theta}_\infty}(\pmb{x})}$. Seeing as this correction term is intractable, Kim \etal \cite{kim2023refining} train a neural discriminator $d_{\pmb{\phi}_*}$ at all noise levels $t$ and use it to estimate the correction term $\pmb{c}_{\pmb{\theta}_\infty}$:
\begin{equation}
    \pmb{c}_{\pmb{\theta}_\infty}(\pmb{x},t) \approx \pmb{c}_{\pmb{\phi}_\infty}(\pmb{x},t) = \nabla \log \frac{d_{\pmb{\phi}}(\pmb{x},t)}{1-d_{\pmb{\phi}}(\pmb{x},t)}
\end{equation}
Thus, they define the discriminator guided reverse SDE as:
\begin{equation}
    d\pmb{x} = [\pmb{f}(\pmb{x}, t) - g(t)^2  (\pmb{s}_{\pmb{\theta}_\infty} + \pmb{c}_{\pmb{\phi}})(\pmb{x},t)] dt + g(t)d \bar{\pmb{w}}.
\end{equation}

The Exposure Bias problem has been uncovered in traditional DDPMs, DDIMs and other diffusion models \cite{ning2023a, ning2023b, li2023alleviating} but not yet in the Discriminator Guidance approach.

\subsection{Prediction Error leads to Exposure Bias}
The exposure bias problem in diffusion models, highlighted by Ning \etal \cite{ning2023a}, refers to the mismatch between the model's input data during the training and the inference phase. During training, the ground truth training sample $\pmb{x}_t$ is available to the model, with the training distribution being $q(\pmb{x}_t|\pmb{x}_0)$. In the inference phase, the model can only rely on the previously generated sample, $\hat{\pmb{x}}_t$. The sampling distribution can be denoted as $q(\hat{\pmb{x}}_t|\pmb{x}_{t+1}, \pmb{x}_{\pmb{\theta}}^{t+1})$, where $\pmb{x}_{\pmb{\theta}}^{t+1}$ is the prediction the model makes for $\pmb{x}_0$ given $\pmb{x}_{t+1}$, using Eq. \ref{xt_x0_cond}, following notation by Ning \etal \cite{ning2023b}. This results in a discrepancy between $\pmb{\epsilon}_{\pmb{\theta}}(\pmb{x}_t)$ and $\pmb{\epsilon}_{\pmb{\theta}}(\hat{\pmb{x}}_t)$.

 Xiao \etal \cite{xiao2021tackling} observed that the sampling distribution $p_{\pmb{\theta}}(\pmb{x}_{t-1}|\pmb{x}_t)$ is parameterized as:
 \begin{equation}
     p_{\pmb{\theta}}(\pmb{x}_{t-1}|\pmb{x}_t) = q(\pmb{x}_{t-1}|\pmb{x}_t, \pmb{x}^t_{\pmb{\theta}})
 \end{equation}
 where $\pmb{x}^t_{\pmb{\theta}}$ represents the predicted $\pmb{x}_0$. The sampling process involves predicting $\pmb{\epsilon}$ using $\pmb{\epsilon_\theta}(\pmb{x}_t, t)$ and deriving the estimation $\pmb{x}^t_{\pmb{\theta}}$ for $\pmb{x}_0$ using Eq. \ref{xt_x0_cond}. Then, $\pmb{x}_{t-1}$ is generated based on the ground truth posterior $q(\pmb{x}_{t-1}|\pmb{x}_t, \pmb{x}_0)$, by replacing $\pmb{x}_0$ with $\pmb{x}^t_{\pmb{\theta}}$. However, $q(\pmb{x}_{t-1}|\pmb{x}_t, \pmb{x}_0) = q(\pmb{x}_{t-1}|\pmb{x}_t, \pmb{x}^t_{\pmb{\theta}})$ holds only if $\pmb{x}^t_{\pmb{\theta}}=\pmb{x}_0$. In practice, $q(\pmb{x}_{t-1}|\pmb{x}_t, \pmb{x}_0) \neq q(\pmb{x}_{t-1}|\pmb{x}_t, \pmb{x}^t_{\pmb{\theta}})$, as the network makes prediction errors when estimating $\pmb{x}_0$ and, as a result, $q(\pmb{x}_{t-1}|\pmb{x}_t, \pmb{x}^t_{\pmb{\theta}})$ does not have the same variance as $q(\pmb{x}_{t-1}|\pmb{x}_t, \pmb{x}_0)$.

 Ning \etal \cite{ning2023b} analytically calculate the discrepancy between the training and sampling distribution in DDPMs. They model $\pmb{x}^t_{\pmb{\theta}}$ as $p_{\pmb{\theta}}(\pmb{x}_0|\pmb{x}_t)$ and approximate it by a Gaussian distribution, following Bao \etal \cite{bao2022a, bao2021analytic}. 
 \begin{equation}
     \pmb{x}^t_{\pmb{\theta}} = \pmb{x}_0+e_t\pmb{\epsilon}_0
 \end{equation}
 where $e_t$ is the standard deviation of $\pmb{x}^t_{\pmb{\theta}}$ and $\pmb{\epsilon}_0 \sim {\cal N}(\pmb{0}, \pmb{I})$.
 Their findings are summarized in Table \ref{tab:prediction_var}. It becomes clear that the sampling distribution's variance is always larger than that of the training distribution by a factor of $(\frac{\sqrt{\bar\alpha_t}\beta_{t+1}}{1-\bar\alpha_{t+1}}e_{t+1})^2$. It must be noted that this is the prediction error produced in a single reverse diffusion step. During sampling, the errors accumulate across steps, resulting in the Exposure Bias problem.

 \begin{table}[h]
     \centering
     \begin{tabular}{@{}lcc@{}}
         \toprule
         & Mean & Variance \\
         \midrule
         $q(\pmb{x}_t|\pmb{x}_0)$ & $\sqrt{\bar\alpha_t}\pmb{x}_0$ & $(1-\bar\alpha_t)\pmb{I}$\\
         $q(\hat{\pmb{x}}_t|\pmb{x}_{t+1}, \pmb{x}_{\pmb{\theta}}^{t+1})$ & $\sqrt{\bar\alpha_t}\pmb{x}_0$ & $(1-\bar\alpha_t + (\frac{\sqrt{\bar\alpha_t}\beta_{t+1}}{1-\bar\alpha_{t+1}}e_{t+1})^2)\pmb{I}$\\
         \bottomrule
     \end{tabular}
     \caption{Mean and variance of $q(\pmb{x}_t|\pmb{x}_0)$ and $q(\hat{\pmb{x}}_t|\pmb{x}_{t+1}, \pmb{x}_{\pmb{\theta}}^{t+1})$}
     \label{tab:prediction_var}
 \end{table}

 \begin{figure}[h]
  \centering
   \includegraphics[width=\linewidth]{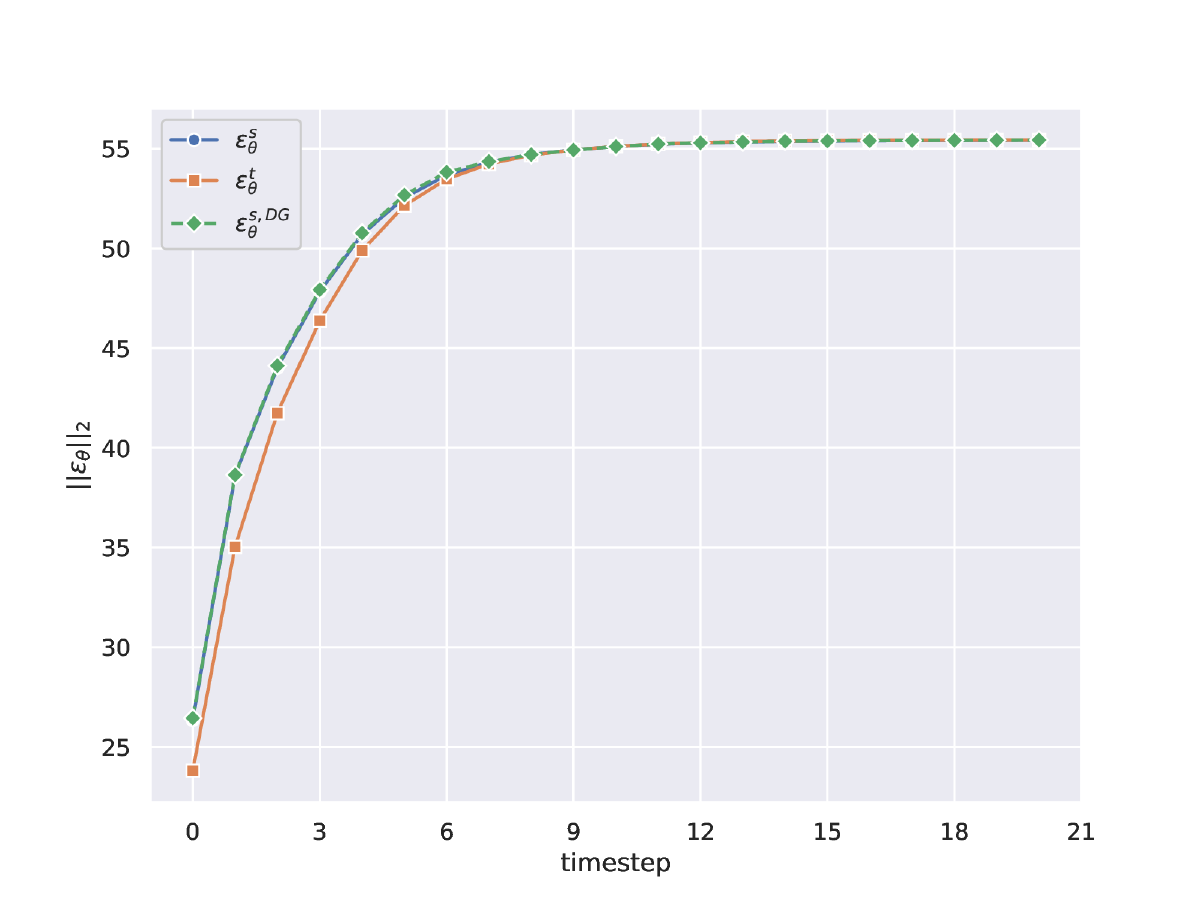}

   \caption{
   EDM model, Euler 1st order solver. $L2$-norm  of $\pmb{\epsilon}_{\pmb{\theta}}(\cdot)$ during 21-step sampling, sampling with DG and training.
   Statistical $L2$-norm was calculated using 50k samples at each timestep.
   Sampling is from right to left.}
   \label{fig:euler_l2}
\end{figure}

 \subsection{Epsilon Scaling}
 \label{sec:epsilon_scaling}
 Ning \etal \cite{ning2023b} propose scaling down the predicted noise factor $\pmb{\epsilon}^s_{\pmb{\theta}}$ (where $s$ denotes the noise factor predicted in the sampling stage) by a factor $\lambda_t$ at time step $t$ as a way to reduce Exposure Bias. Their approach is based on the assumption that the accuracy of the prediction $\pmb{\epsilon}^s_{\pmb{\theta}}$ can be enhanced if we are able to shift the input $(\hat{\pmb{x}}_t, t)$ away from the unreliable vector field (depicted as the orange curve in Fig. \ref{fig:euler_l2} and \ref{fig:heun_l2}) and towards the dependable vector field (represented by the green curve in Fig. \ref{fig:euler_l2} and \ref{fig:heun_l2}).

 Their approach is rooted in the following observation: $\pmb{\epsilon}^s_{\pmb{\theta}}$ and $\pmb{\epsilon}^t_{\pmb{\theta}}$ (where $t$ denotes to the noise factor predicted in the training stage) both originate from the same input $\pmb{x}_T \sim {\cal N} (0, \pmb{I})$ at time step $t = T$. However, starting from time step $T - 1$, $\hat{\pmb{x}}_t$ (the input for $\pmb{\epsilon}^s_{\pmb{\theta}}$) begins to deviate from $\pmb{x}_t$ (the input for $\pmb{\epsilon}^t_{\pmb{\theta}}$) due to the $\pmb{\epsilon}_{\pmb{\theta}}(\cdot)$ error made in the previous time step. This iterative process continues throughout the sampling chain, leading to exposure bias. Therefore, we can bring $\hat{\pmb{x}}_t$ closer to $\pmb{x}_t$ by reducing the overestimated magnitude of $\pmb{\epsilon}^s_{\pmb{\theta}}$. Their sampling method only differs from Eq. \ref{eq:mu_sample} in the $\lambda_t$ term and can be expressed as:
 \begin{equation}
     \mu_{\pmb{\theta}}(\pmb{x}_t,t) = \frac{1}{\sqrt{a_t}}\left(\pmb{x}_t-\frac{\beta_t}{\sqrt{1-\bar a_t}} \frac{\pmb{\epsilon_\theta} (\pmb{x}_t,t)}{\lambda_t}\right)
 \end{equation}
 As a result, epsilon scaling is a plug-in method which requires no retraining or fine-tuning of the original score model and adds no overhead computational cost.
\section{Method}
\subsection{Quantifying Exposure Bias}
As Table \ref{tab:prediction_var} illustrates, the $\pmb{x}_t$ seen by the network during training differs from the $\hat{\pmb{x}}_t$ seen by the network during sampling. This discrepancy leads to a drift between the noise prediction made during training, $\pmb{\epsilon}^t_{\pmb{\theta}}$, and the noise prediction made during sampling, $\pmb{\epsilon}^s_{\pmb{\theta}}$. 
Following Ning \etal \cite{ning2023b}, we choose to measure the sampling drift at each timestep as the difference between $\pmb{\epsilon}^t_{\pmb{\theta}}$ and $\pmb{\epsilon}^s_{\pmb{\theta}}$. However, since the ground truth of $\pmb{\epsilon}^s_{\pmb{\theta}}$ is intractable in the sampling phase, we use the $L2$-norm to quantify the exposure bias \cite{ning2023b}.

\begin{figure}[h]
  \centering
   \includegraphics[width=\linewidth]{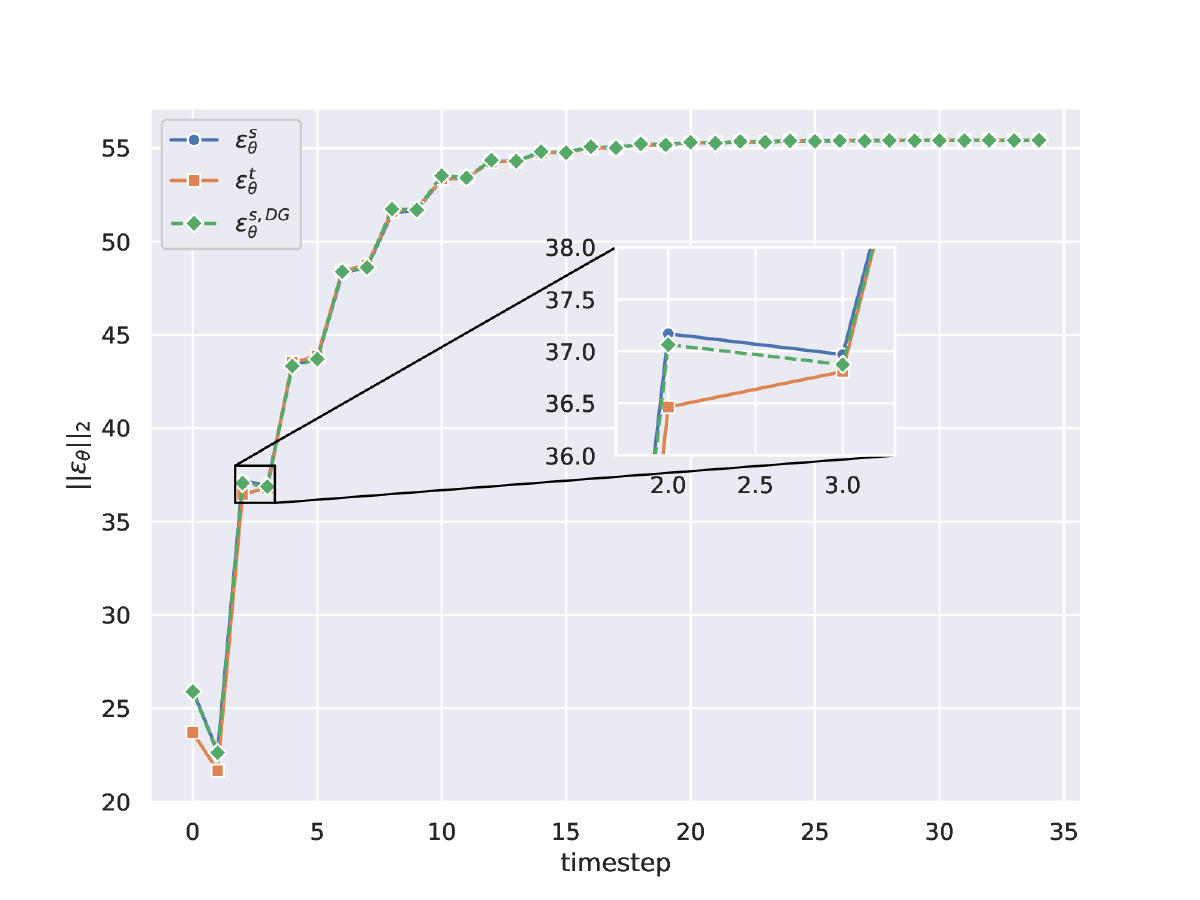}
   \caption{EDM model, Heun 2nd order solver. $L2$-norm  of $\pmb{\epsilon}_{\pmb{\theta}}(\cdot)$ during 35-step sampling, sampling with DG and training.
   Statistical $L2$-norm was calculated using 50k samples at each timestep.
   Sampling is from right to left.}
   \label{fig:heun_l2}
\end{figure}

In Fig. \ref{fig:euler_l2} and Fig. \ref{fig:heun_l2} we plot the $L2$-norm of $\pmb{\epsilon}^t_{\pmb{\theta}}$,  $\pmb{\epsilon}^s_{\pmb{\theta}}$ and $\pmb{\epsilon}^{s, DG}_{\pmb{\theta}}$ using the Euler, as well as the Heun ODE solver. $\pmb{\epsilon}^s_{\pmb{\theta}}$ and $\pmb{\epsilon}^{s, DG}_{\pmb{\theta}}$ refer to the noise prediction in the vanilla EDM model and in the discriminator guided EDM-G++ model, respectively. We observe that in the case of the Euler solver, the $L2$-norm of $\pmb{\epsilon}^{s, DG}_{\pmb{\theta}}$ is larger than that of $\pmb{\epsilon}^t_{\pmb{\theta}}$ and nearly coincides with the $L2$-norm of $\pmb{\epsilon}^s_{\pmb{\theta}}$. In the case of the Heun 2nd order solver, the difference between the two norms is smaller, however, the the $L2$-norm of $\pmb{\epsilon}^{s, DG}_{\pmb{\theta}}$ is, once again, larger than that of $\pmb{\epsilon}^t_{\pmb{\theta}}$ and closer to that of $\pmb{\epsilon}^s_{\pmb{\theta}}$. This means that the correction term offered by discriminator guidance does not alleviate the sampling procedure of its collected exposure bias. Instead, the prediction errors accumulate and the learnt vector field $\pmb{\epsilon}^{s, DG}_{\pmb{\theta}}$ deviates from the desired sampling trajectory.

\subsection{Proposed Framework}
We use the EDM model, proposed by Karras \etal \cite{karras2022edm}, as a score estimator due to the detailed way in which it was designed. Through careful network design and fine-tuning of hyperparameters, EDM achieves a substantial quality enhancement, reducing the FID score on the CIFAR-10 dataset to 1.97, demonstrating notable progress at the time. Moreover, seeing as Exposure Bias exhibits a strong correlation with FID score \cite{ning2023b}, we assume that in comparison to other networks, the vanilla EDM model demonstrates a reduced accumulation of exposure bias.

When it comes to the discriminator network, we follow the setup demonstrated by Kim \etal \cite{kim2023refining}. The discriminator network is comprised of the encoders of two U-Net structures. The first U-Net encoder is frozen and the second one is fine-tuned, to accelerate training. We leave the exploration of other discriminator architectures, such as Vision Transformers, as future work.

In Sec. \ref{sec:related_work}, we mention different approaches which seek to reduce the impact of Exposure Bias on the sampling trajectory. Ning \etal \cite{ning2023a} suggest introducing an extra noise factor at each step during the training to mitigate the discrepancy between training and inference. However, this method proves cumbersome as it requires retraining the model from scratch. Li \etal \cite{li2023alleviating} explore how manipulation of the time step during the reverse generation process can trick the model into reducing the Exposure Bias issue. This method is also difficult to implement as exploring the entire space of possible combinations is inconvenient as it requires costly experimentation to produce noteworthy results. The latest method to reduce Exposure Bias, introduced by Ning \etal \cite{ning2023b}, known as Epsilon Scaling, is our selected approach. It is a training free, plug-in method, which has proven effective in reducing Exposure Bias and significantly improving the FID score across a range of diffusion models (ADM \cite{dhariwal2021diffusion}, DDIM \cite{song2020denoising}, EDM \cite{karras2022edm}, LDM \cite{rombach2022high}). We present the main notion of Epsilon Scaling in Sec. \ref{sec:epsilon_scaling}.

Although the network output of EDM is the score function $\pmb{s}_{\pmb{\theta}}$, not $\pmb{\epsilon}$, the noise factor $\pmb{\epsilon}$ can easily be extracted at each sampling step and used to apply Epsilon Scaling \cite{ning2023b}.

When it comes to designing the scaling schedule $\lambda_t$, Ning \etal \cite{ning2023b} propose that the the term $\lambda_t$ should be a linear function $\lambda_t = kt + b$ where $k$, $b$ are constants. They also observe that the longer the sampling step, the smaller the $k$ that should be used. Thus, they suggest a uniform schedule $\lambda_t (k=0)$ to facilitate practicality and simplify the exploration of the $b$ parameter. In our experiments, we confirm that the use of a linear $\lambda_t$ can provide similar or sub-optimal results, compared to the uniform schedule $\lambda_t=b$ and explore the constant scaling factor more extensively.
\section{Results}
\subsection{Euler Solver}
\begin{figure}[h]
    \centering
    \includegraphics[width=\linewidth]{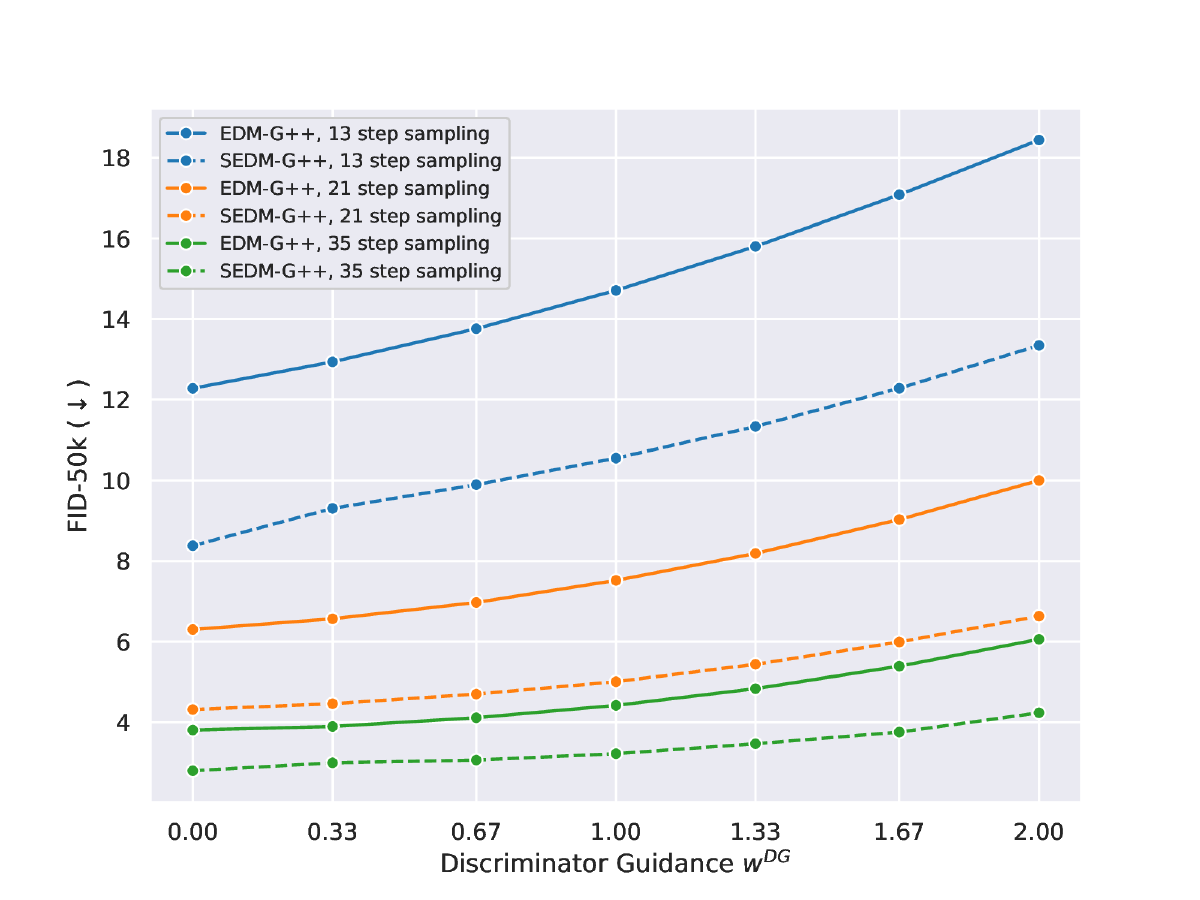}
    \caption{FID-50k ablation study on DG weight (Euler Solver).}
    \label{fig:euler_results}
\end{figure}

We firstly present our results using the Euler 1st order ODE solver. When it comes to the Discriminator Guidance method (Kim \etal \cite{kim2023refining}), we identify a noteworthy omission. Namely, the authors do not report performance of the EDM-G++ model using the Euler ODE solver. We have calculated the FID score of EDM-G++ on the CIFAR-10 dataset using various numbers of timesteps and Discriminator Guidance weight, $w^{DG}$ and we present our results in Fig. \ref{fig:euler_results}. 

Notably, the EDM-G++ model, which is guided by a discriminator, exhibits a noteworthy decrease in sample quality when compared to the baseline EDM model. This decline intensifies as the weight assigned to the discriminator's correction term, $w^{DG}$, increases. This outcome is intriguing in light of the fact that the inclusion of discriminator guidance led to a substantial reduction in the FID score using the Heun 2nd order ODE solver. It is plausible to surmise that the reduced FID score in the case of the 1st order solver may be attributed to the absence of a corrective step in the sampling process. The enhancement of sample quality in the EDM model \cite{karras2022edm} is markedly facilitated by the inclusion of a corrective step in the ODE solver. This is a crucial factor contributing to the widespread adoption of the Heun ODE solver in recent research, as it consistently delivers superior performance \cite{karras2022edm}.

Nevertheless, the epsilon scaling method demonstrates its efficacy in the context of discriminator-guided diffusion models as well. SEDM-G++ successfully reduces the FID score across different numbers of total timesteps and $w^{DG}$ values when utilizing the Euler ODE solver, as compared to the EDM-G++ and EDM baselines. Remarkably, SEDM-G++ narrows the performance gap between EDM-G++ using a 21-step Euler solver and a 35-step Euler solver, with the performance of SEDM-G++ using a 21-step Euler solver closely approaching that of the baseline EDM-G++ model using a 35-step Euler solver. This results in a significant reduction in computational and time requirements during inference, without compromising sample quality to a considerable extent.

\begin{figure}[h]
  \centering
  \begin{subfigure}{0.45\linewidth}
    \includegraphics[width=\linewidth]{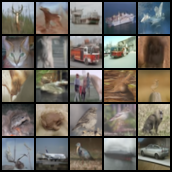}
    \caption{EDM-G++}
    \label{fig:samples_1}
  \end{subfigure}\hfill
  \begin{subfigure}{0.45\linewidth}
    \includegraphics[width=\linewidth]{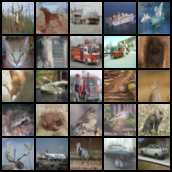}
    \caption{SEDM-G++}
    \label{fig:samples_2}
  \end{subfigure}
  \caption{Uncoordinated samples from the EDM-G++ baseline and our proposed SEDM-G++.}
  \label{fig:euler_samples}
\end{figure}

In Figure \ref{fig:euler_samples}, we present uncoordinated samples derived from the EDM-G++ baseline and our proposed SEDM-G++. Apart from the evident improvement in the FID score, our method yields noticeable qualitative enhancements in the generated samples. For example, the sample located in the third row and third column, as well as the sample in the fifth row and fourth column, exhibit a substantial enhancement compared to the baseline. In Figure \ref{fig:samples_1}, the shapes appear blurry and obscure, making it challenging to discern the content of the images. Conversely, in Figure \ref{fig:samples_2}, SEDM-G++ produces images with clear, well-defined shapes and vivid colors, rendering the identity of the depicted objects readily discernible.

\subsection{Heun Solver}
\begin{figure}[h]
    \centering
    \includegraphics[width=0.8\linewidth]{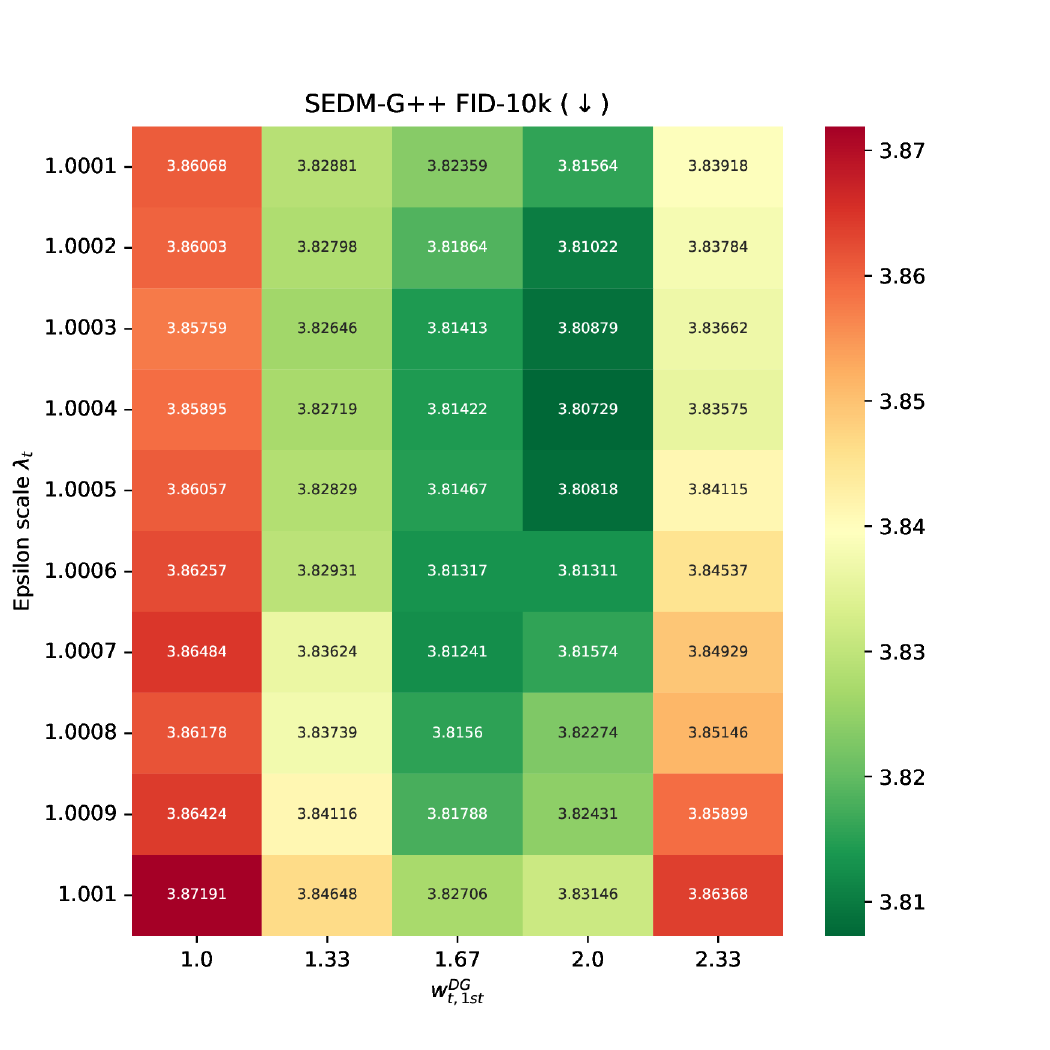}
    \caption{FID-10k ablation study on DG weight and scaling factor (Heun Solver).}
    \label{fig:ablation_10k}
\end{figure}

We further explore the performance of SEDM-G++ using the Heun 2nd order ODE solver. We conducted an ablation study on the relation between the weight attributed to discriminator guidance in the 1st order step of the Heun solver, namely $w^{DG}_{t,1st}$, and the epsilon scaling factor $\lambda_t=b$. Based on the work of Kim \etal \cite{kim2023refining}, we set the 2nd order discriminator guidance weight, namely $w^{DG}_{t,2nd}$ equal to zero for all tests. This choice offers optimal sample quality and requires requires fewer computational resources, as the number of calls to the discriminator network is practically halved by omission in the 2nd order corrective steps. In order to reduce the computational needs of our study, rather than generating a total of 50k samples, we generate 10k samples for each setting and derive the FID-10k score, which suffices for the purpose of parameter optimization \cite{ning2023b}. The results are presented in Fig. \ref{fig:ablation_10k}.

\begin{figure}[h]
    \centering
    \includegraphics[width=\linewidth]{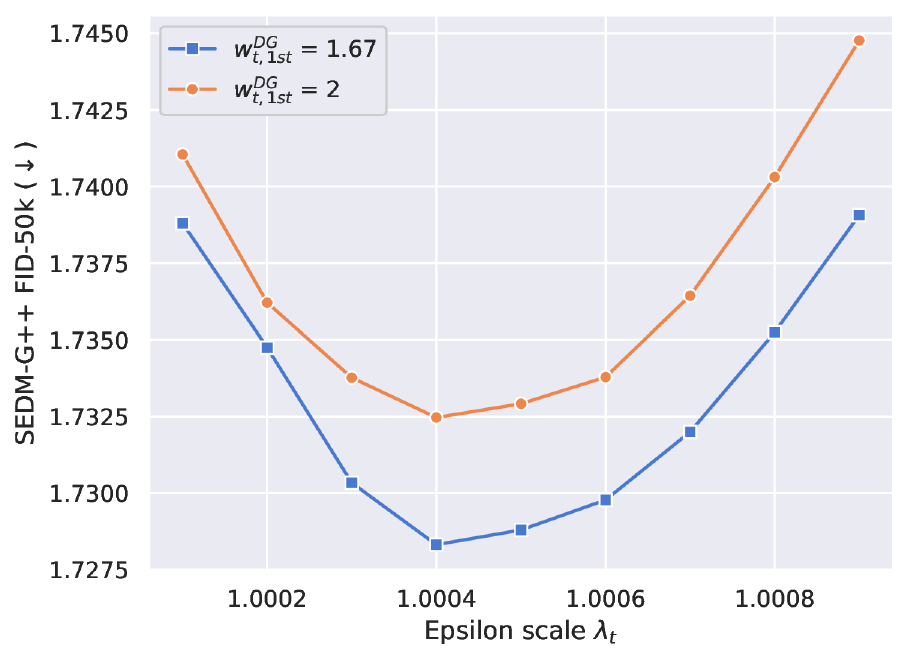}
    \caption{FID-50k ablation study on best performing DG weight values (Heun Solver).}
    \label{fig:ablation_50k}
\end{figure}

Our observations indicate that the most effective discriminator guidance weight values are 1.67 and 2. Notably, when comparing these two values, we also note that the optimal epsilon scaling value $\lambda_t$ decreases as the discriminator's weight coefficient increases. We further delve into the performance of the best-performing $w^{DG}_{t,1st}$ values through a comprehensive study, by generating 50k samples for each setting and utilizing the proper FID-50k score to compare. We present our results in Fig. \ref{fig:ablation_50k}.

Our proposed SEDM-G++ outperforms the current state-of-the-art in unconditional CIFAR-10 image generation, by achieving an FID score of 1.73. The optimal hypermarameters used are $\lambda_t=1.0004$, $w^{DG}_{t,1st}=1.67$ and $w^{DG}_{t,2nd}=0$. A comprehensive comparison between SEDM-G++ and other prominent diffusion models is provided in Table \ref{tab:comparison}. Given that our approach is based in the EDM model \cite{karras2022edm}, it maintains a low Number of Function Evaluations (NFE) at 35 network calls per batch. This figure is significant as it directly relates to the computational cost associated with the sampling process.

\begin{table}[h]
    \centering
    \begin{tabular}{c|c|c}
        Model&  NFE($\downarrow$) &FID($\downarrow$)\\
        \toprule
        \small VDM (Kingma \etal, 2021 \cite{kingma2021variational}) & 1000 & 7.41 \\
        \small DDPM (Ho \etal, 2020 \cite{ho2020ddpm}) & 1000 & 3.17 \\
        \small iDDPM (Nichol \& Dhariwal, 2021 \cite{nichol2021improved}) & 1000 & 2.90 \\
        \small Soft Truncation (Kim \etal, 2022 \cite{kim2022soft}) & 2000 & 2.47 \\
        \small INDM (Kim \etal, 2022 \cite{kim2022maximum}) & 2000 & 2.28 \\
        \small CLD-SGM (Dockhorn \etal, 2022 \cite{dockhorn2021score}) & 312 & 2.25 \\
        \small NCSN++ (Song \etal, 2020 \cite{song2021sde}) & 2000 & 2.20 \\
        \small LSGM (Vahdat \etal, 2021 \cite{vahdat2021score}) & 138 & 2.10 \\
        \small EDM (Karras \etal, 2022 \cite{karras2022edm}) & \textbf{35} & 1.97 \\
        \small EDM-G++ (Kim \etal, 2023 \cite{kim2023refining}) & \textbf{35} & 1.77 \\
        \small \textbf{SEDM-G++ (ours)} & \textbf{35} & \textbf{1.73} \\
        \hline
        \multicolumn{3}{p{0.9\linewidth}}{\scriptsize \textbf{Note:} Following the work of Karras \etal \cite{karras2022edm}, we calculate the FID for different seeds and report the minimum. Kim \etal \cite{kim2023refining} use random seeds for FID calculation. Manually calculating the FID of EDM-G++ results in an FID of 1.75.}
    \end{tabular}
    \caption{FID-50k performance comparison on unconditional CIFAR-10 image generation.}
    \label{tab:comparison}
\end{table}

An intriguing observation pertains to the fact that the FID gain achieved through Epsilon Scaling in the Euler sampler is more pronounced compared to the case of the Heun sampler. This is in line with the observations of Ning \etal \cite{ning2023b}, who attribute this phenomenon to two key factors. Firstly, higher-order ODE solvers, such as Heun solvers, entail a lower level of truncation error in contrast to Euler 1st order solvers. Secondly, the corrective steps integrated into the Heun solver serve to mitigate exposure bias by readjusting the drifted sampling trajectory back to the precise vector field. This is also evident in Fig. \ref{fig:euler_l2} and \ref{fig:heun_l2}. In the case of the Heun solver, any prediction error (the root cause of exposure bias) incurred during each Euler step is rectified during the subsequent correction step (Fig. \ref{fig:heun_l2}), leading to a reduction in exposure bias. This exposure bias perspective offers a comprehensive explanation for the superior performance of the Heun solver in diffusion models.
\section{Conclusion}

We explore the effectiveness of Discriminator Guidance in addressing exposure bias accumulation during the sampling process in diffusion models. Our findings reveal that, despite notable improvements in sample quality, Discriminator Guidance falls short in mitigating exposure bias. In response, we introduce SEDM-G++, a novel approach that integrates a modified sampling technique, incorporating both Discriminator Guidance and Epsilon Scaling. Applying this method to the pre-trained EDM model, we demonstrate its consistent ability to enhance sample quality while reducing exposure bias. This improvement is observed across various ODE solvers, a range of numbers of timesteps employed, and different hyperparameter settings. Our proposed approach outperforms the current state-of-the-art, achieving an FID score of 1.73 on the unconditional CIFAR-10 dataset.

\paragraph{Future Work} In the ongoing expansion of this research, we intend to evaluate the applicability of this framework across a broader range of datasets, with past research suggesting a robust generalization to various domains \cite{ning2023b, kim2023refining}. Exploring combinations of different strategies to minimize exposure bias \cite{ning2023a, ning2023b, li2023alleviating} has the potential to improve sample quality and also serves as a direction for future research. Furthermore, exploring the integration of a different state-of-the-art architecture for the discriminator part of the framework holds the potential for performance improvement.
{
    \small
    \bibliographystyle{ieeenat_fullname}
    \bibliography{main}
}


\end{document}